%% file: iclr2023_conference.tex
\documentclass{article} 
\usepackage{iclr2023_conference,times}

\input{math_commands.tex}

\usepackage{tikz}
\usepackage{hyperref}
\usepackage{url}
\usepackage{natbib}

\usepackage{multirow}
\usepackage{longtable}

\title{DeepEpiSolver: Unravelling Inverse problems in Covid, HIV, Ebola and Disease Transmission}

\iclrfinalcopy

\author{Ritam Majumdar, Shirish Karande \& Lovekesh Vig\\
Tata Consultancy Services Research, India\\
\texttt{\{ritam.majumdar,shirish.karande,lovekesh.vig\}@tcs.com} \\
}

\begin{document}

\maketitle

\begin{abstract}

The spread of many infectious diseases is modeled using variants of the SIR compartmental model, which is a coupled differential equation. The coefficients of the SIR model determine the spread trajectories of disease, on whose basis proactive measures can be taken. Hence, the coefficient estimates must be both fast and accurate. Shaier et al. in \cite{DBLP:journals/corr/abs-2110-05445} used Physics Informed Neural Networks (PINNs) to estimate the parameters of the SIR model. There are two drawbacks to this approach. First, the training time for PINNs is high, with certain diseases taking close to 90 hrs to train. Second, PINNs don't generalize for a new SIDR trajectory, and learning its corresponding SIR parameters requires retraining the PINN from scratch. In this work, we aim to eliminate both of these drawbacks. We generate a dataset between the parameters of ODE and the spread trajectories by solving the forward problem for a large distribution of parameters using the LSODA algorithm. We then use a neural network to learn the mapping between spread trajectories and coefficients of SIDR in an offline manner. This allows us to learn the parameters of a new spread trajectory without having to retrain, enabling generalization at test time. We observe a speed-up of 3-4 orders of magnitude with accuracy comparable to that of PINNs for 11 highly infectious diseases. Further finetuning of neural network inferred ODE coefficients using PINN further leads to 2-3 orders improvement of estimated coefficients.
\end{abstract}

\section{Introduction}

In the past century alone, mankind has been suffering from various deadly diseases, with Spanish flu (1918-1920), Hong Kong Flu (1968-1969), Ebola Virus (1976-ongoing), Middle East Respiratory Syndrome (MERS) (2012-ongoing), Ebola Virus (2014), Severe acute respiratory syndrome coronavirus-2 (SARS-CoV-2) just to name a few. The recent outbreak of SARS-CoV-2 has led to a few million infections across the globe, and the importance of mathematical modeling of these infectious diseases has never been more important. In the early stages of the pandemic, mathematical models helped public health officials understand the potential impact of the disease and predict how quickly it could spread. This information was used to make informed decisions about which mitigation measures were necessary and which were most effective.
Additionally, mathematical models were also used to evaluate the effectiveness of different vaccination strategies and make predictions about the future course of the pandemic. This information was critical in helping countries prioritize vaccine distribution and allocate resources.  

One of the earliest methods for predicting the spread of infectious disease was formulated by the Kermack–McKendrick epidemic model \cite{doi:10.1098/rspa.1927.0118}. In this model, the population under consideration is divided into three classes, susceptible (S), infective (I), and removed (R). Over the years, the SIR model has been improved to account for hospitalizations, latent dynamics, posthumous transmissions, etc to improve the predictions of mathematical models like COVID \cite{NBERw27128}, Anthrax\cite{article}, HIV\cite{PERELSON199381}, Zika\cite{Gao_2016}, Smallpox\cite{Gani2001TransmissionPO}, Tuberculosis\cite{article1}, Pneumonia\cite{article2}, Ebola\cite{10.2307/4617539}, Dengue\cite{article3}, Polio\cite{BUNIMOVICHMENDRAZITSKY2005302} and Measles\cite{Bolker1995SpacePA}. Shaier et al. in \cite{DBLP:journals/corr/abs-2110-05445} used Physics-informed Neural Networks \cite{RAISSI2019686} to predict the coefficients of mathematical models in \cite{NBERw27128,article,PERELSON199381,Gao_2016,Gani2001TransmissionPO,article1,article2,10.2307/4617539,article3,BUNIMOVICHMENDRAZITSKY2005302,Bolker1995SpacePA} for one specific SIR trajectory instance. The PINNs are trained in an unsupervised manner and use a physics-informed loss to solve the inverse problem of coefficient estimation of differential equations. The physics-informed loss, however, has a complex loss landscape \cite{article4} and struggles to converge quickly, leading to larger training durations, with some diseases requiring around 90 hrs to train. Furthermore, PINNs must be retrained for a new trajectory estimation, as the underlying parameters governing the trajectory differ. In this work, we circumvent both these problems. We generate a dataset wherein we solve the computationally cheaper forward problem multiple times to generate spread trajectories, given the parameters of the SIR model. Then we use a neural network to learn the inverse mapping between trajectories and parameters. We use this trained neural network to estimate the parameters of the SIR model for a new test task, avoiding retraining. We observe a speed-up of 3-4 orders of magnitude, while parameter estimations are on par with PINNs. Further finetuning of neural network inferred parameters using PINNs outperform
PINNs from scratch by 2-3 orders of magnitude at 25 times lower train time.

\section{DeepEpiSolver Methodology}

\begin{figure}
    \centering
    \includegraphics[height=10cm]{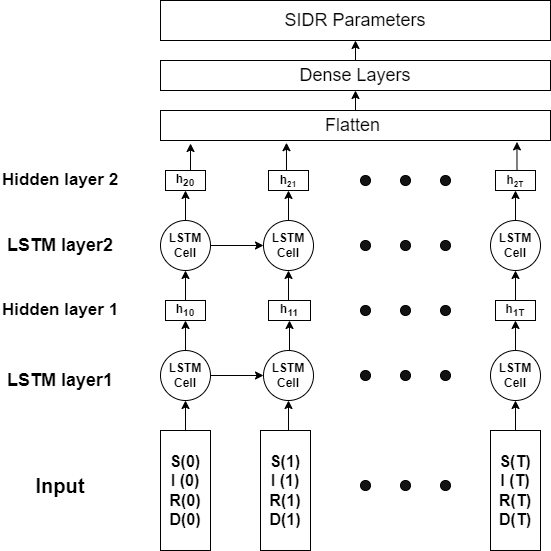}
    \caption{Two layered LSTM architecture}
    \label{fig:LSTM}
\end{figure}

We first generate the dataset. We solve the forward problem of generating spread trajectories, given the parameters of the disease ODEs using the LSODA \cite{HindmarshSep2005} algorithm. The LSODA algorithm is computationally very cheap and just takes 2-3 seconds for simulating the trajectories per example. In the next step, we use a neural network to learn the mapping between spread trajectory and parameters. As the spread trajectories have temporal variations, a multi-layered LSTM \cite{8478136} is the most natural choice of Neural Network. Figure \ref{fig:LSTM} represents our multi-layered LSTM architecture. At every time step, we pass the spread population $[S(t)$,$I(t)$,$R(t)$,$D(t)]$, where $t\in[0,T]$ as input, which is fed to the multilayered LSTM. The final hidden output layers of LSTM are flattened and passed through dense layers to obtain the final estimations of the parameters of the ODE. We further finetune the inferred LSTM-obtained parameters using a Physics-informed Neural Network using the same training settings as described in \cite{DBLP:journals/corr/abs-2110-05445}.

\section{Experiments}

Table \ref{tab:DE_information} consists of the information on the differential equations governing the diseases. We normalize our inputs (trajectories) before passing through the neural network. We use a multi-layer LSTM architecture with 2 layers and 256 hidden neurons to learn the mapping between the trajectories and the parameters, followed by flattening and 2 dense layers with 64 neurons. We consider 9000 training examples, 500 validation, and 500 test examples for every disease. Table \ref{tab:parameterwise_error} consists of the ranges of each parameter to be estimated for every disease. We use an Adam Optimizer for 60k epochs with multiplicative decay of 0.1 every 20k epochs with an initial learning rate of $1e^{-3}$. We use relative percentage-L2-error as our metric for parameters with non-zero ground-truth value, and mean-average-error for parameters having zero as their true value. All experiments were conducted on NVIDIA P100 GPU with 16 GB GPU Memory and 1.32 GHz GPU Memory clock using Pytorch framework.

\section{Covid-19}

In this section, we elaborate upon the mathematical model of Covid-19 and discuss the impact of its coefficients. The mathematical model governing the spread of Covid-19 \cite{NBERw27128} is modeled as follows:

\begin{subequations}
\begin{align}
d S / d t &= -(\alpha / N) S I\\
d I / d t &= (\alpha / N) S I-\beta I-\gamma I\\
d D / d t &= \gamma I\\
d R / d t &= \beta I
\end{align}
\end{subequations}

\begin{figure}[h]
    \centering
    \includegraphics[width=\textwidth, height=8cm]{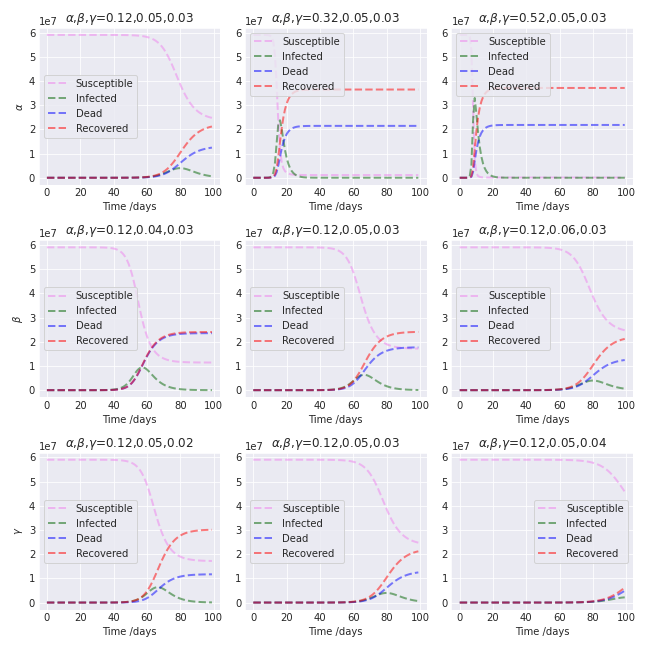}
    \caption{Influence of SIDR parameters on Covid trajectories}
    \label{fig:Covid}
\end{figure}

Here S, I, D, and R stand for susceptible, infected, dead, and recovered populations respectively. $\alpha\in[0.12,0.52]$,$\beta\in[0.04,0.06]$ and $\gamma\in[0.02,0.04]$ are the parameters that control the trajectories. Figure \ref{fig:Covid} provides a visualization of how the trajectories are influenced by changes in the range of parameters, with the first, second, and third rows representing changes w.r.t $\alpha$,$\beta$,$\gamma$ respectively. We observe, as the value of $\alpha$ increases, the infection peaks in a lesser number of days, and reaches a higher peak. An increase in $\beta$ and $\gamma$ leads to lowering and delaying the infection and death trajectory peak. $\beta$ and $\gamma$ has an opposite effect as $\alpha$.

\section{Observations}

\begin{table}
    \centering
    \begin{tabular}{ccccccccc}
        &\multicolumn{4}{c}{\% Relative L2-error}&\multicolumn{4}{c}{Time statistics}\\
        \hline
        &\multicolumn{2}{c}{LSTM}&\multicolumn{2}{c}{PINN-Finetune}&\multicolumn{2}{c}{LSTM}&\multicolumn{2}{c}{PINN}\\
        &Mean&Std-dev&Mean&Std-dev&Training&Test&Training&Finetuning\\
        \hline
        Covid &0.7366&0.067&$1.71e^{-3}$&$3.59e^{-5}$&0.85 hrs&2 s&20 mins& 11 mins\\
        HIV &0.1691&0.0124&$1.19e^{-3}$&$6.67e^{-5}$&1.14 hrs&2 s&22 hrs&1.25 hrs\\
        Small Pox&3.2057&1.2611&$2.44e^{-3}$&$4.47e^{-5}$&1.25 hrs&2 s&14 hrs&0.75 hrs\\
        Tuberculosis &2.9033&0.2271&$2.62e^{-3}$&$2.76e^{-5}$&1.16 hrs&2 s&12 hrs&0.75 hrs\\
        Pneumonia &2.2597&0.3917&$1.05e^{-3}$&$8.17e^{-5}$&1.34 hrs&2 s&41 hrs&1.25 hrs\\
        Dengue &1.5690&0.1609&$8.41e^{-4}$&$6.24e^{-5}$&1.62 hrs&2 s&33 hrs&1.45 hrs\\
        Ebola &2.4076&0.1741&$2.69e^{-3}$&$6.38e^{-5}$&1.46 hrs&2 s&58 hrs&1.5 hrs\\
        Anthrax &2.0815&0.7015&$5.18e^{-4}$&$1.43e^{-5}$&2.21 hrs&2 s&91 hrs&1.5 hrs\\
        Polio &0.2601&0.025&$3.88e^{-3}$&$2.15^{-5}$&1.75 hrs&2 s&66 hrs&1.5 hrs\\
        Measles&2.4145&0.355&$5.57e^{-3}$&$1.09e^{-5}$&1.82 hrs&2 s&28 hrs&2 hrs\\
        Zika &0.2416&0.0375&$4.29e^{-3}$&$8.33e^{-5}$&2.06 hrs&2 s&13 hrs&2.25 hrs\\
        \hline
    \end{tabular}
    \label{tab:Results}
\end{table}

Table \ref{tab:Results} summarizes the results of our experiments. We provide two results: LSTM-inferred parameters (Column 2,3) and fine-tuned parameters by PINNs using LSTM-inferred parameters (Column 4,5) as initialization. The statistics of mean and standard deviation of relative-L2-error for every disease are across all test tasks and all parameter coefficients of interest. We observe that PINN-based finetuning improves the error estimates by 2-3 orders, almost learning the true parameters perfectly, and also reduces the standard deviation by 4 orders. HIV, Small Pox has the smallest and largest LSTM mean error, while Anthrax and Measles have the smallest and largest PINN-finetuned mean errors. Columns 6-9 provide the time statistics. We observe the training time for LSTM to generalize for 9000 train tasks is approximately 18.64 times faster than the train time for PINN for just 1 task. Furthermore, as LSTMs generalize the mapping between trajectories and SIR parameters, they can directly infer the parameters of the SIR differential equation for a new set of spread trajectories without retraining, unlike PINNs. Thus, LSTMs take just 2 s to infer for a new task. Additionally, the convergence time of PINN finetuning is on average 25 times lower than PINNs from scratch, indicating the importance of better initialization to faster convergence of PINNs.

In table \ref{tab:parameterwise_error}, for every disease, we provide the range of the parameter of the differential equation and its corresponding parameter-wise $\%$ relative-L2-errors for LSTM inference and PINN finetuning. For LSTM, we observe across all diseases, the parameters which higher magnitude values (in the range of $1e^{0}$ or higher) have a lower mean and standard deviation in contrast to parameters with low magnitude values ($1e^{-1}$ and lower). For a fair comparison with existing literature, we compare the performance of LSTM at test-time viz-a-viz with PINNs for the spread trajectories in \cite{DBLP:journals/corr/abs-2110-05445}, and compile its results in table \ref{tab:Comparison} and Figure \ref{fig:All}. We observe, apart from Covid, HIV, and Zika, LSTMs provide better estimates than PINNs for a majority of the parameters. Even on the parameters where LSTMs are worse (all diseases including Covid, HIV, and Zika), the estimates are not too far off from PINNs and have a negligible difference from the true trajectories, as observed in figure \ref{fig:All}. Finetuning the parameters using PINNs on top of LSTM inference further improves the parameter estimates, as observed by 2-3 orders lower error over both LSTM inferred parameters and PINNs trained from scratch.  

\section{Conclusion}

In this work, we show NNs are capable of learning the mapping between disease spread trajectories and the coefficients of the differential equation governing them. NNs can generalize for a new system of trajectories and predict the ODE coefficients accurately without requiring re-training. This eliminates the two drawbacks suffered by PINNs: Longer time and inability to generalize for a new task. Finetuning ODE parameters using PINNs leads to an even better estimate of ODE parameters at 25 times lesser train time than PINNs with the random initialization, further improving the training cost. Quicker and more accurate estimation of governing parameter coefficients allows for proactive measures like hospitalization, vaccination, or lockdowns, which can significantly help reduce human casualties. 

\bibliography{iclr2023_conference}
\bibliographystyle{iclr2023_conference}

\appendix
\section{Appendix}

Comparison of LSTM with PINN in literature task
\begin{table}
    \centering
    \begin{tabular}{|c|l|}
    \hline
       \multirow{3}{*}{HIV}   & $d T / d t=s-\mu_T T+r T\left(1-\left((T+I) / T_{\max }\right)-k_1 V T\right)$ \\
& $d I / d t=k_1^{\prime} V T-\mu_I I$ \\
& $d V / d t=N \mu_b I-k_1 V T-\mu_V V$ \\
      \hline   
      \multirow{8}{*}{Small-Pox} & $d S / d t=\chi_1\left(1-{ }_1\right) C i-\beta(\phi+\rho-\phi \rho) S I$ \\
& $d E n / d t=\beta \phi(1-\rho) S I-\alpha E n$ \\
& $d E i / d t=\beta \phi \rho S I-\left(\chi_{12}+\alpha\left(1-{ }_2\right)\right) E i$ \\
& $d C i / d t=\beta \rho(1-\phi) S I-\chi_1 C i$ \\
& $d I / d t=\alpha(1-\theta) E n-(\theta+\gamma) I$ \\
& $d Q / d t=\alpha\left(1-{ }_2\right) E i+\theta(\alpha E n+I)-\chi_2 Q$ \\
& $d U / d t=\gamma I+\chi_2 Q$ \\
& $d V / d t=\chi_1\left({ }_2 E i+{ }_1 C i\right)$ \\
\hline
\multirow{4}{*}{Tuberculosis}& $d S / d t=\delta-\beta c S I / N-\mu S$ \\
& $d L / d t=\beta c S I / N-\left(\mu+k+r_1\right) L+\beta^{\prime} c T / N$ \\
& $d I / d t=k L-(\mu+d) I-r_2 I$ \\
& $d T / d t=r_1 L+r_2 I-\beta^{\prime} c T / N-\mu T$\\
\hline
\multirow{5}{*}{Pneumonia}&$ d S / d t=(1-p) \pi+\phi V+\delta R-(\mu+\lambda+\theta) S$ \\
& $d V / d t=p \pi+\theta S-(\mu+\lambda+\phi) V $\\
& $d C / d t=\rho \lambda S+\rho \lambda V+(1-q) \eta I-(\mu+\beta+\chi) C $\\
& $d I / d t=(1-\rho) \lambda S+(1-\rho) \lambda V+\chi C-(\mu+\alpha+\eta) I$ \\
& $d R / d t=\beta C+q \eta I-(\mu+\delta) R$\\
\hline
\multirow{7}{*}{Dengue}&$d S b / d t  =\pi_{b}-\lambda_{b} S b-\mu_{b} S b$ \\
&$d E b / d t  =\lambda_{b} S b-\left(\sigma_{b} \mu_{b}\right) E b$ \\
&$d I b / d t  =\sigma_{b} E b-\left(\tau_{b}+\mu_{b}+\delta_{b}\right) I b$ \\
&$d R b / d t  =\tau_{b} I b-\mu_{b} R b$ \\
&$d S v / d t  =\pi_{v}-\delta_{v} S v-\mu_{v} S v $\\
&$d E v / d t  =\delta_{v} S v-\left(\sigma_{v}+\mu_{v}\right) E v$ \\
&$d I v / d t  =\sigma_{v} E v-\left(\mu_{v}+\delta_{v}\right) I v$ \\
\hline
\multirow{6}{*}{Ebola}&$d S / d t=-1 / N\left(\beta_{1} S I+\beta_{b} S H+\beta_{f} S F\right)$\\
&$d E / d t=1 / N\left(\beta_{1} S I+\beta_{b} S H+\beta_{f} S F\right)-\alpha E$\\
&$d I / d t=\alpha E-\left(\gamma_{b} \theta_{1}+\gamma_{i}\left(1-\theta_{1}\right)\left(1-\delta_{1}\right)+\gamma_{d}\left(1-\theta_{1}\right) \delta_{1}\right) I$\\
&$d H / d t=\gamma_{b} \theta_{1} I-\left(\gamma_{d} b \delta_{2}+\gamma_{i} b\left(1-\delta_{2}\right)\right) H$\\
&$d F / d t=\gamma_{d}\left(1-\theta_{1}\right) \delta_{1} I+\gamma_{d} b \delta_{2} H-\gamma_{f} F$\\
&$d R / d t=\gamma_{i}\left(1-\theta_{1}\right)\left(1-\delta_{1}\right) I+\gamma_{i} b\left(1-\delta_{2}\right) H+\gamma_{f} F$\\
\hline
\multirow{4}{*}{Anthrax}& $d S / d t=r(S+I)(1-(S+I) / K)-\eta_a A S-\eta_c S C-\eta_i(S I) /(S+I)-\mu S+\tau I$ \\
& $d I / d t=\eta_a A S+\eta_c S C+\left(\eta_i(S I) /(S+I)-(\gamma+\mu+\tau)\right) I $\\
& $d A / d t=-\sigma A+\beta C$ \\
& $d C / d t=(\gamma+\mu) I-\delta(S+I) C-\kappa C$\\
\hline
\multirow{6}{*}{Polio}& $d S c / d t=\mu N-\left(\alpha+\mu+\left(\beta_{c c} / N c\right) I c+\left(\beta_{c a} / N c\right) I a\right) S c$ \\
& $d S a / d t=\alpha S c-\left(\mu+\left(\beta_{a a} / N a\right) I a+\left(\beta_{a c} / N a\right) I c\right) S a$ \\
& $d I c / d t=\left(\left(\beta_{c c} / N c\right) I c+\left(\beta_{c a} / N c\right) I a\right) S c-\left(\gamma_c+\alpha+\mu\right) I c$ \\
& $d I a / d t=\left(\left(\beta_{a c} / N a\right) I c+\left(\beta_{a a} / N a\right) I a\right) S a-\left(\gamma_a+\mu\right) I a+\alpha I c$ \\
& $d R c / d t=\gamma_c I c-\mu R c-\alpha R c$ \\
& $d R a / d t=\gamma_a I a-\mu R a+\alpha R c$ \\
\hline
\multirow{3}{*}{Measles}& $d S / d t=\mu(N-S)-(\beta S I) / N$ \\
& $d E / d t=(\beta S I) / N-(\mu \sigma) E$ \\
& $d I / d t=\sigma E-(\mu+\gamma) I$ \\
\hline
\multirow{9}{*}{Zika}&$d S b / d t=-a b(I v / N b) S b-\beta\left(\left(\kappa E b+I b_1+\tau I b_2\right) / N b\right) S b$ \\
& $d E b / d t=\theta\left(-a b(I v / N b) S b-\beta\left(\left(\kappa E b+I b_1+\tau I b_2\right) / N b\right) S b\right)-V_b E b$ \\
& $d I b_1 / d t=V_b E b-\gamma_{b 1} I b_1$ \\
& $d I b_2 / d t=\gamma_{b 1} I b_1-\gamma_{b 2} I b_2$ \\
& $d A b / d t=(1-\theta)\left(a b(I v / N b) S b-\beta\left(\left(\kappa E b+I b_1+\tau I b_2\right) / N b\right) S b\right)-\gamma_b A b$ \\
& $d R b / d t=\gamma_{b 2} I b_2+\gamma_b A b$ \\
& $d S v / d t=\mu_v N v-a c\left(\left(\eta E b+I b_1\right) / N b\right) S v-\mu_v S v$ \\
& $d E v / d t=a c\left(\left(\eta E b+I b_1\right) / N b\right)-\left(V_v+\mu_v\right) E v$ \\
& $d I v / d t=V_v E v-\mu_v I v$ \\
\hline
    \end{tabular}
    \caption{Differential Equations governing the diseases}
    \label{tab:DE_information}
\end{table}

\begin{longtable}{ccccccccc}
&&&\multicolumn{3}{c}{Predicted Parameter}&\multicolumn{3}{c}{\% Relative L2-Error}\\
\hline
&& Actual & PINN & LSTM & Finetune & PINN & LSTM & Finetune \\
\hline
\multirow{3}{*}{Covid}&$\alpha$ & $0.191$ & $0.193$ & {0.193}& 0.191 &$1.151$ & {0.837} & $1.1e^{-4}$\\
&$\beta$ & $0.05$ & $0.0501$ & 0.050 & {0.049} &$0.2$ &{0.400} & $2.4e^{-4}$\\
&$\gamma$ & $0.029$ & $0.029$ & {0.029} & 0.029 &$1.02$ &{1.701} & $4.7e^{-4}$\\
\hline
\multirow{10}{*}{HIV}&$s$ & 10  & ${10.001}$ &10.05 & 10.001 &  ${0.007}$ &0.500&$2.26e^{-3}$ \\
&$\mu_T$ & $0.02$  & ${0.021}$ &0.0198 & 0.0200 & ${3.812}$&4.633&$1.59e^{-2}$ \\
&$\mu_I$ & $0.26$  & $0.261$ & 0.260 & {0.257}& $0.489$ &{1.154}&$3.31e^{-2}$\\
&$\mu_b$ & $0.24$  & $0.242$ & {0.239} & 0.240& $0.727$ &{0.417}&$2.23e^{-3}$\\
&$\mu_V$ & $2.4$ & $2.419$ & {2.407} & 2.400 & $0.829$ &{0.292}&$1.15^{-3}$\\
&$r$ & $0.03$  & $0.031$ & {0.030} & 0.03&  $2.016$ &{1.333}&$7.74e^{-3}$\\
&$N$ & 250  & ${249.703}$ & 248.85 & 250.001& ${0.118}$ &0.460&$2.56e^{-3}$ \\
&$T_{\max }$ & 1500 & $1.506e^{3}$ & $1.491e^{3}$ & 1499.96&  ${0.436}$ &0.563&$1.49e^{-3}$\\
&$k_1$ & $2.4e^{-5}$ & {$2.46e^{-5}$} & 2.19e-5 & $2.4e^{-5}$ & ${2.448}$ &8.75&$9.17e^{-3}$\\
&$k_1^{\prime}$ & $2e^{-5}$ & $2.03e^{-5}$ & 1.94e-5& 1.99e-5 & {$1.599$} &3.00&$3.75e^{-3}$\\
\hline
\multirow{8}{*}{Small Pox}&$\chi_1$ & $0.06$  & ${0.0554}$ &$0.0539$ & 0.5994 & ${7.7222}$ &11.31&$1.16e^{-2}$\\
&$\chi_2$ & $0.04$  & $0.0380$ &{0.0407} & 0.040 & $4.924$ &{1.749}&$5.95e^{-3}$\\
&1 & $0.975$  & ${0.9839}$ & 0.9625 & 0.975 & ${0.909}$ &1.282&$2.69e^{-3}$\\
&2 & $0.3$  & ${0.2841}$ & 0.2951 & 0.3 & ${5.285}$ & 1.633&$7.71e^{-3}$\\
&$\rho$ & $0.975$  & ${0.9759}$ & 0.971 & 0.975 & ${0.088}$ &0.574&$1.88e^{-2}$\\
&$\theta$ & $0.95$  & $0.9050$ & {0.981} & 0.95& $4.737$ &{3.263}&$4.61e^{-2}$\\
&$\alpha$ & $0.068$  & $0.0626$  & {0.065} & 0.068 & $8.549$&{4.558}&$3.85e^{-3}$\\
&$\gamma$ & $0.11$  & $0.1034$ & {0.111} & 0.11 & $10.859$ &{1.181}&$7.24^{-2}$\\
\hline
\multirow{8}{*}{Tuberculosis}&$\delta$ & 500  & ${509.4698}$ & 483.48& 500.011&${1.8587}$& 3.3039&$1.55e^{-3}$\\
&$\beta$ & 13 &  $12.5441$  & {12.87}& 13.0004& $3.6341$&{2.514}&$1.24e^{-3}$\\
&$c$ & 1  & $1.0405$ & {0.984}& 1.0003 & $3.8952$ &{2.16}&$1.11e^{-3}$\\
&$\mu$ & $0.143$  & $0.1474$ & {0.141}& 0.143 &$3.0142$ &{1.398}&$7.84e^{-4}$\\
&$k$ & $0.5$  & $0.5396$ & {0.4985}& 0.5001& $7.3433$ &{0.300}&$1.62e^{-4}$\\
&$r_1$ & 2 & ${1.9892}$ & 2.034& 1.9996 & ${0.5424}$ &1.69&$5.16e^{-3}$\\
&$r_2$ & 1 & $1.1243$ & {1.0431}& 1.0013& $11.0583$ &{4.31}&$2.97e^{-3}$\\
&$\beta^{\prime}$ & 13  & $13.7384$ & {12.99}& 13.001& $5.3746$ & {0.076}&$8.68e^{-5}$\\
&$d$ & 0  & $-0.0421$ & {0.012}& $2.4e^{-3}$& $0.0421$  & {0.012}&$2.4e^{-3}$\\
\hline
&$\pi$ & $0.01$  & $0.0098$ & {0.0101}& 0.010004 & $2.0032$ & {1.000}&$2.22e^{-4}$\\
&$\lambda$ & $0.1$  & $0.0990$ & {0.0994}& 0.1& $0.9622$ &{0.600}&$2.22e^{-4}$\\
&$k$ & $0.5$  & ${0.5025}$ & 0.5086& 0.5& ${0.5083}$ &1.72&$5.19e^{-3}$\\
&$\chi$& $0.002$  & $0.0022$ & {0.00195}& 0.00201& $11.6847$ &{2.500}&$4.76e^{-3}$\\
\multirow{5}{*}{Pneumonia}&$\tau$ & $0.89$  & $0.8912$ & {0.8858}& 0.0890& $0.1309$ &{0.4917}&$5.19e^{-3}$\\
&$\phi$ & ${0.0025}$  & ${0.0027}$ & 0.0024& $0.00250$&$7.4859$ &4.00&$3.56e^{-3}$\\
&$\chi$ & $0.001$  & ${0.0011}$ & 0.0009& $0.001$&${6.7374}$ &10.00&$2.29e^{-2}$\\
&$p$ & $0.2$  & $0.2033$ & {0.1985}& 0.2001&  $1.6372$ &{0.7500}&$7.77e^{-5}$\\
&$\theta$ & $0.008$  & $0.0084$ & {0.0081}& $0.008$&1$4.8891$ &{1.249}&$5.95e^{-3}$\\
&$\mu$ & $0.01$  & $0.0092$ & {0.0103}& 0.010&$8.4471$ &{2.999}&$3.88e^{-3}$\\
&$\alpha$ & $0.057$  & ${0.0570}$ & 0.0556& 0.057& ${0.0005}$ &2.456&$1.07e^{-2}$\\
&$\rho$ & $0.05$  & ${0.0508}$ & 0.0489& 0.0499 & ${1.5242}$ &2.200&$1.07e^{-2}$\\
&$\beta$ & $0.0115$  & $0.0122$ & {0.0113}& 0.011501 & $5.8243$ &{1.739}&$1.55e^{-3}$\\
&$\eta$ & $0.2$  & ${0.2023}$ & 0.1958& 0.200& ${1.1407}$  &2.100&$3.55e^{-3}$\\
&$q$ & $0.5$  & $0.4960$ & {0.5037}& 0.5004& $0.8003$ &{0.7400}&$1.15e^{-4}$\\
&$\delta$ & $0.1$  & ${0.1038}$ & 0.1106& 0.1003& ${3.7502}$ &10.59&$1.35e^{-2}$\\
\hline
\multirow{11}{*}{Dengue}&$\pi_b$ & 10  & $9.9317$ & {9.6514}& 9.9995 & $0.6832$ &{0.3538}&$3.19e^{-4}$\\
&$\pi_v$ & 30  & $29.8542$ & {30.042}& 30.001& $0.4859$ &{0.1400}&$1.71e^{-4}$\\
&$\lambda_b$ & $0.055$  & $0.0552$ & {0.0551}& 0.055& $0.2696$ &{0.18}&$1.96e^{-5}$\\
&$\lambda_v$ & $0.05$  & $0.0506$ & {0.0495}& 00499& $1.2876$ &{1.00}&$5.56e^{-4}$\\
&$\delta_b$ & $0.99$  & $0.9643$ & {1.003}& 0.9902&$2.5967$ &{1.31}&$1.02e^{-3}$\\
&$\delta_v$ & $0.057$  & ${0.0567}$ & 0.0558& 0.057& ${0.5294}$ &2.1052&$4.28e^{-3}$\\
&$\mu_b$ & $0.0195$  & $0.0194$ & {0.01952}& 0.0195& $0.3835$ &{0.1025}&$6.24e^{-5}$\\
&$\mu_v$ & $0.016$  & ${0.0159}$ & 0.0163& 0.016 & ${0.8796}$ &1.874&$4.74e^{-3}$\\
&$\sigma_b$ & $0.53$  & ${0.5372}$ & 0.5219& 0.5301& ${1.3567}$ &1.528&$2.65e^{-4}$\\
&$\sigma_v$ & $0.2$  & $0.1989$ & {0.2004}& 0.2& $0.5483$ &{0.199}&$1.96e^{-5}$\\
&$\tau_b$ & $0.1$  & $0.0902$ & {0.1013}& 0.1001&$9.7723$ &{1.299}&$5.27e^{-4}$\\
\hline
\multirow{13}{*}{Ebola}&$\beta_1$ & $3.532$ & $3.5589$ & {3.5474}& 2.532& $0.7622$ &{0.6317}&$5.59e^{-4}$\\
&$\beta_b$ & $0.012$  & $0.0129$ & {0.0118}& 0.0120& $7.8143$ &{5.5134}&$6.31e^{-4}$\\
&$\beta_f$ & $0.462$  & $0.4638$ & {0.4629}& 0.46201& $0.3976$ &{0.2784}&$4.41e^{-3}$\\
&$\alpha$ & $1 / 12$  & $0.0866$ & {0.0866} & 0.0833& $3.9320$ &{3.9320}&$2.66e^{-3}$\\
&$\gamma_b$ & $1 / 4.2$  & $0.2471$ & {0.2392}& 0.2380& $3.7853$ &{3.6124}&$7.71e^{-4}$\\
&$\theta_1$ & $0.65$  & ${0.6523}$ & 0.6534& 0.6497& ${0.3477}$ &0.5624&$5.81e^{-5}$\\
&$\gamma_i$ & $0.1$  & $0.0904$ & {0.0906}& 0.0999& $9.6403$ &{9.400}&$1.41e^{-3}$\\
&$\delta_1$ & $0.47$  & ${0.4712}$ & 0.4750& 0.4699& ${0.2565}$ &1.0638&$8.58e^{-3}$\\
&$\gamma_d$ & $1 / 8$  & ${0.1205}$ & 0.1199& 0.12499& ${3.6124}$ &4.079&$2.36e^{-3}$\\
&$\delta_2$ & $0.42$  & $0.4247$ & {0.4169}& 0.42003& $1.1244$ &{0.7380}&$6.74e^{-4}$\\
&$\gamma_f$ & $0.5$  & $0.5196$ & {0.5074}& 0.4998& $3.9246$ &{1.479}&$5.71e^{-4}$\\
&$\gamma_{i b}$ & $0.082$  & ${0.0811}$ & 0.0809& 0.08207& ${1.0932}$ &1.3416&$2.19e^{-4}$\\
&$\gamma_{d b}$ & $0.07$  & $0.0710$ & {0.0704}& 0.070& $0.7563$ &{0.5714}&$2.00e^{-4}$\\
\hline
\multirow{12}{*}{Anthrax}&$\mathrm{r}$ & $1 / 300$  & ${0.0034}$ & {0.0032} & 0.00332& ${1.2043}$&{1.2043}&$3.27e^{-3}$\\
&$\mu$ & $1 / 600$  & ${0.0017}$ & {0.0017} & 0.001667 & ${0.5754}$  & {0.5754}&$3.27e^{-3}$ \\
&$\kappa$ & $0.1$  & $0.1025$ & {0.0999} & 0.100& $2.5423$ &{0.100}&$4.16e^{-5}$\\
&$\eta_a$ & $0.5$  & $0.5035$ & {0.4979}& 0.5001& $0.7022$ &{0.420}&$2.19e^{-4}$\\
&$\eta_c$ & $0.1$  & $0.1024$ & {0.1014}& 0.1003&$2.4492$ &{1.400}&$3.55e^{-3}$\\
&$\eta_i$ & $0.01$  & $0.0106$ & {0.0101}& 0.0100& $6.0459$ &{0.999}&$4.39e^{-3}$\\
&$\tau$ & $0.1$  & $0.0976$ & {0.0995}& 0.0999&$2.4492$ &{0.500}&$7.72e^{-4}$\\
&$\gamma$ & $1 / 7$  & $0.1444$ & {0.1429}& 0.14285& $1.0542$ &{0.030}&$6.16e^{-3}$\\
&$\delta$ & $1 / 20$  & $0.0512$ & {0.0505}& 0.0500& $2.3508$ & {1.002}&$8.04e^{-3}$\\
&$K$ & 100  & ${100.6391}$ & 98.73& 100.005 &${0.6391}$ &1.269&$5.18e^{-3}$\\
&$\beta$ & $0.02$  & $0.0021$ & {0.0199}& 0.0199& $6.5466$ &{0.499}&$6.14e^{-3}$\\
&$\sigma$ & $0.1$  & $0.1051$ & {0.1006}& 0.1004& $5.1029$  & {0.6014}&$7.44e^{-4}$\\
\hline
\multirow{8}{*}{Polio}&$\mu$ & $0.02$  & 0.0199& ${0.0200}$ & 0.020& ${0.0}$ &0.4995&$5.21e^{-4}$\\
&$\alpha$ & $0.5$  & $0.5018$ & {0.5007}& 0.5002& $0.36$  & {0.140} &$2.41e^{-4}$\\
&$\gamma_a$ & 18  & $18.0246$ & {17.9751}& 18.003& $0.4168$ &{0.135}&$1.29e^{-4}$\\
&$\gamma_c$ & 36  & $36.0701$ & {35.9541}& 36.007 &$0.3587$ &{0.1388}&$1.01e^{-4}$\\
&$\beta_{a a}$ & 40  & $40.2510$ & {39.9814}& 40.005& $0.6275$ &{0.050}&$1.37e^{-5}$\\
&$\beta_{c c}$ & 90  & $90.6050$ & {89.4736}& 89.994& $0.6722$ &{0.5888}&$8.61e^{-5}$\\
&$\beta_{a c}$ & 0  & ${0.0002}$ & 0.0019&$1.7e^{-5}$&${0.0002}$  & 0.0019&$1.7e^{-5}$\\
&$\beta_{c a}$ & 0  & ${0.0004}$ & {-0.0004}& $2.1e^{-6}$& ${0.0004}$ & {0.0004}& $2.1e^{-6}$\\
\hline
\multirow{4}{*}{Measles}&$\mu$ & $0.02$  & $0.0225$ & {0.0198}& 0.0200 & $12.704$ &{0.999}&$3.65e^{-3}$\\
&$\beta_1$ & $0.28$  & ${0.2700}$ & 0.2633& 0.2801&${3.5704}$ &5.9642&$4.79e^{-3}$\\
&$\gamma$ & 100  & $97.0001$ & {100.59}& 100.003& $2.9999$  & {0.5900}&$7.87e^{-4}$\\
&$\sigma$ & $35.84$  & $34.7127$ & {36.18}& 36.007& $3.1453$ &{0.9486}&$4.53e^{-3}$\\
\hline
& $\mathrm{a}$ & $0.5$ & 0.5116 & ${0.4997}$ &0.5001 & ${0.0588}$ &2.32&$3.89e^{-2}$\\
&$\mathrm{b}$ & $0.4$  & ${0.4033}$ & 0.3965& 0.4& ${0.8297}$ &1.6865&$7.36e^{-3}$\\
&$\mathrm{c}$ & $0.5$  & $0.5015$ & {0.4986}& 0.5001& $0.3086$ &{0.2800}&$7.66e^{-3}$\\
&$\eta$ & $0.1$  & ${0.0999}$ & 0.0963& 0.1007& ${0.0687}$ &3.6036&$2.98e^{-2}$\\
&$\beta$ & $0.05$  & ${0.0498}$ & {0.0502}& 0.05& ${0.4098}$  & {0.4098}&$7.31e^{-3}$\\
&$\kappa$ & $0.6$  & ${0.6033}$ & {0.5967}& 0.6004& ${0.5486}$  & {0.5486}&$2.99e^{-2}$\\
&$\tau$ & $0.3$  & $0.2902$ & {0.2969}& 0.3& $3.2565$ &{2.3087}&$1.72e^{-2}$\\
\multirow{3}{*}{Zika}&$\theta$ & 18  & ${17.9669}$ & 18.821& 18.006& ${0.1838}$ &4.5611&$1.04e^{-2}$\\
&$\mathrm{m}$ & 5 & $5.2937$ & {4.9621}& 5.0009& $5.8748$ &{0.7600}&$5.99e^{-3}$\\
&$V_b$ & $1 / 5$  & ${0.1996}$ & 0.2014& 0.2003&${0.1798}$ &0.6999&$4.85e^{-3}$\\
&$V_v$ & 10  & ${10.0170}$ & 9.965& 10.001& ${0.1700}$ &0.3049&$2.77e^{-4}$\\
&$\gamma_{b 1}$ & $1 / 5$  & $0.1991$ & {0.2009}& 0.2006& $0.4651$ &{0.449}&$1.96e^{-3}$\\
&$\gamma_{b 2}$ & $1 / 20$  & $0.0504$ & {0.0497}& 0.05& $0.7261$ &{0.6003}&$2.17e^{-3}$\\
&$\gamma_b$ & $1 / 7$ & $0.1406$ & {0.1412}& 0.1427& $1.5967$ &{1.16}&$8.94e^{-3}$\\
&$\mu_v$ & $1 / 14$ & ${0.0723}$ & 0.0725& 0.0714& ${1.1806}$ &1.499&$5.25e^{-3}$\\
\hline
\label{tab:Comparison}
\end{longtable}

Parameter-wise statistics for inferred-LSTM and finetuned-PINN

\begin{longtable}{ccccccc}
&&&\multicolumn{2}{c}{LSTM-inference}&\multicolumn{2}{c}{finetuned-PINN}\\
\hline
&Parameter & Range & Mean & Std. Dev & Mean & Std. Dev \\
\hline
\multirow{3}{*}{Covid}&$\alpha$  & $(0.12,0.52)$ & 1.04 & 0.15 & $1.69e^{-3}$ & $3.64e^{-5}$  \\
&$\beta$  & $(0.04,0.06)$ & 0.18 & 0.03 & $1.65e^{-3}$ & $3.45e^{-5}$\\
&$\gamma$  & $(0.02,0.04)$ & 0.99 & 0.02 & $1.77e^{-3}$ & $3.68e^{-5}$\\
\hline
\multirow{10}{*}{HIV}&$s$  & $(5,15)$ & 0.0109 & 0.008& $1.17e^{-3}$ & $6.67e^{-5}$\\
&$\mu_T$  & $(0.005,0.045)$ & 2.5614 & 0.3917& $1.18e^{-3}$ & $6.64e^{-5}$\\
&$\mu_I$  & $(0.05,0.45)$ & 0.5564 & 0.0318& $1.25e^{-3}$ & $6.64e^{-5}$\\
&$\mu_b$  & $(0.05,0.45)$ & 0.7548 & 0.0147& $1.22e^{-3}$ & $6.68e^{-5}$\\
&$\mu_V$  & $(2,3)$ & 0.8767 & 0.0435& $1.20e^{-3}$ & $6.75e^{-5}$\\
&$r$  & $(0.01,0.05)$ & 2.1614 & 0.1947& $1.16e^{-3}$ & $6.71e^{-5}$\\
&$N$  & $(225,275)$ & 0.0878 & 0.0314& $1.18e^{-3}$ & $6.60e^{-5}$\\
&$T_{\max }$  & $(1400,1600)$ & 0.4519 & 0.041& $1.23e^{-3}$ & $6.80e^{-5}$\\
&$k_1$  & $\left(2e^{-5}, 3e^{-5}\right)$ & 2.031 & 0.414& $1.24e^{-3}$ & $6.59e^{-5}$\\
&$k_1^{\prime}$  & $\left(1.5e^{-5}, 2.5e^{-5}\right)$ & 1.6514 & 0.0056& $1.11e^{-3}$ & $6.61e^{-5}$\\
\hline
\multirow{8}{*}{Small Pox}&$\chi_1$  & $(0.04,0.08)$ & 4.4817 & 1.0814&$2.44e^{-3}$&$4.46e^{-5}$\\
&$\chi_2$  & $(0.02,0.06)$ & 2.136 & 0.465&$2.49e^{-3}$&$4.58e^{-5}$\\
&1  & $(0.75,1.25)$ & 0.775 & 0.064&$2.39e^{-3}$&$4.39e^{-5}$\\
&2  & $(0.1,0.5)$ & 3.3954 & 1.0314&$2.41e^{-3}$&$4.35e^{-5}$\\
&$\rho$  & $(0.9,1.1)$ & 0.619 & 0.155&$2.47e^{-3}$&$4.52e^{-5}$\\
&$\theta$  & $(0.5,1.5)$ & 2.235 & 1.16&$2.44e^{-3}$&$4.46e^{-5}$\\
&$\alpha$  & $(0.02,0.1)$ & 4.175 & 2.19 &$2.43e^{-3}$&$4.39e^{-5}$\\
&$\gamma$  & $(0.05,0.15)$ & 4.523 & 2.74 &$2.45e^{-3}$&$4.58e^{-5}$\\
\hline
\multirow{8}{*}{Tuberculosis}&$\delta$  &$(450,550)$ & 1.1936 & 0.043&$2.65e^{-3}$&$2.75e^{-5}$\\
&$\beta$  & $(5, 25)$ & 2.5617 & 0.0716 &$2.60e^{-3}$&$2.77e^{-5}$\\
&$c$   & $(0.3, 1.7)$ & 2.5514 & 0.3926 &$2.63e^{-3}$&$2.76e^{-5}$\\
&$\mu$  &  $(0.08, 0.22)$  & 2.9147 & 0.1007 &$2.63e^{-3}$&$2.77e^{-5}$\\
&$k$ & $(0.05,1.2)$ & 7.0414 & 0.3591 &$2.64e^{-3}$&$2.74e^{-5}$\\
&$r_1$  &  $(0.5,5.5)$ & 0.5115 & 0.0321 &$2.62e^{-3}$&$2.79e^{-5}$\\
&$r_2$  & $(0.4,2.4)$ & 5.3147 & 0.3628 &$2.62e^{-3}$&$2.75e^{-5}$\\
&$\beta^{\prime}$  & $(5,20)$ & 4.0225 & 0.7762 &$2.61e^{-3}$&$2.78e^{-5}$\\
&$d$ & $(-1,1)$ & 0.0184 & 0.0065 &$2.64e^{-3}$&$2.70e^{-5}$\\
\hline
\multirow{16}{*}{Pneumonia}&$\pi$  & $(0.005,0.015)$ & 1.165 & 0.134&$1.05e^{-3}$&$8.18e^{-5}$\\
&$\lambda$  & $(0.05,0.15)$ &  0.6514 & 0.2014&$1.04e^{-3}$&$8.14e^{-5}$\\
&$k$ & $(0.3,0.7)$ & 0.5017 & 0.0091&$1.04e^{-3}$&$8.11e^{-5}$\\
&$\chi$ & $(0.0005,0.0045)$ & 6.6911 & 1.5614&$1.04e^{-3}$&$8.18e^{-5}$\\
&$\tau$  & $(0.6,1.2)$ & 0.0514 & 0.0412 &$1.04e^{-3}$&$8.16e^{-5}$\\
&$\phi$  & $(0.0005,0.0045)$ & 5.218 & 0.631 &$1.07e^{-3}$&$8.15e^{-5}$\\
&$\chi$  & $(0.0005,0.0015)$ & 1.054 & 0.365&$1.06e^{-3}$&$8.21e^{-5}$\\
&$p$ & $(0.1,0.3)$ & 1.555 & 0.008 &$1.06e^{-3}$&$8.14e^{-5}$\\
&$\theta$  & $(0.005,0.0011)$ &  3.871 & 0.514&$1.07e^{-3}$&$8.20e^{-5}$\\
&$\mu$  & $(0.005,0.015)$ & 5.591 & 1.135&$1.04e^{-3}$&$8.13e^{-5}$\\
&$\alpha$  & $(0.04,0.08)$ & 0.0004 & 0.0001&$1.05e^{-3}$&$8.11e^{-5}$\\
&$\rho$  & $(0.01,0.09)$ & 1.835 & 0.3217&$1.04e^{-3}$&$8.17e^{-5}$\\
&$\beta$  & $(0.0085,0.0145)$ & 3.162 & 0.821 &$1.08e^{-3}$&$8.19e^{-5}$\\
&$\eta$  & $(0.1,0.3)$ &  0.756 & 0.026&$1.07e^{-3}$&$8.17e^{-5}$\\
&$q$  & $(0.3,0.7)$ & 0.642 & 0.115 &$1.05e^{-3}$&$8.22e^{-5}$\\
&$\delta$  & $(0.05,0.15)$ & 3.311 & 0.515 &$1.08e^{-3}$&$8.09e^{-5}$\\
\hline
&$\pi_b$  & $(5,15)$ & 0.4334 & 0.1124&$8.40e^{-4}$&$6.24e^{-5}$ \\
&$\pi_v$  & $(25,35)$ & 0.4251 & 0.0514&$8.44e^{-4}$&$6.16e^{-5}$ \\
&$\lambda_b$  & $(0.01,0.09)$ & 0.3014 & 0.0314 &$8.34e^{-4}$&$6.17e^{-5}$\\
&$\lambda_v$  & $(0.01,0.09)$ & 1.1129 & 0.128&$8.38e^{-4}$&$6.26e^{-5}$ \\
&$\delta_b$  & $(0.75,1.25)$ & 2.336 & 0.217 &$8.48e^{-4}$&$6.33e^{-5}$\\
\multirow{4}{*}{Dengue}&$\delta_v$  & $(0.04,0.08)$ & 0.497 & 0.025&$8.37e^{-4}$&$6.29e^{-5}$\\
&$\mu_b$  & $(0.0175,0.0225)$ & 0.333 & 0.117 &$8.51e^{-4}$&$6.27e^{-5}$\\
&$\mu_v$  & $(0.006,0.026)$ & 0.8325 & 0.0469 &$8.43e^{-4}$&$6.22e^{-5}$\\
&$\sigma_b$  & $(0.4,0.8)$ & 1.265 & 0.4112 &$8.36e^{-4}$&$6.26e^{-5}$\\
&$\sigma_v$  & $(0.05,0.45)$ & 0.6317 & 0.072 &$8.39e^{-4}$&$6.20e^{-5}$\\
&$\tau_b$  & $(0.05,0.15)$ & 9.092 & 0.464 &$8.22e^{-4}$&$6.31e^{-5}$\\
\hline
\multirow{13}{*}{Ebola}&$\beta_1$  & $(3,4)$ & 0.6525 & 0.0759& $2.69e^{-3}$& $6.45e^{-5}$ \\
&$\beta_b$  & $(0.005,0.015)$ & 4.165 & 1.114 & $2.65e^{-3}$& $6.39e^{-5}$ \\
&$\beta_f$  & $(0.3,0.7)$ & 0.2256 & 0.1035 & $2.71e^{-3}$& $6.41e^{-5}$ \\
&$\alpha$  & $(0.05,0.15)$ & 3.265 & 0.551& $2.74e^{-3}$& $6.37e^{-5}$ \\
&$\gamma_b$  & $(0.05,0.45)$ & 3.157 & 0.556& $2.67e^{-3}$& $6.32e^{-5}$  \\
&$\theta_1$  & $(0.4,0.8)$ & 0.3911 & 0.393& $2.67e^{-3}$& $6.33e^{-5}$  \\
&$\gamma_i$ & $(0.05,0.15)$ & 8.815 & 0.45 & $2.69e^{-3}$& $6.38e^{-5}$ \\
&$\delta_1$  & $(0.3,0.9)$ & 0.3161 & 0.1014 & $2.70e^{-3}$& $6.38e^{-5}$ \\
&$\gamma_d$  & $(0.1,0.3)$ & 3.321 & 0.469& $2.75e^{-3}$& $6.47e^{-5}$  \\
&$\delta_2$  & $(0.2,0.8)$ & 1.078 & 0.1015 & $2.64e^{-3}$& $6.46e^{-5}$ \\
&$\gamma_f$  & $(0.2,0.8)$ & 3.963 & 0.062 & $2.66e^{-3}$& $6.38e^{-5}$ \\
&$\gamma_{i b}$  & $(0.05,0.15)$ & 1.1374 & 0.0521 & $2.72e^{-3}$& $6.35e^{-5}$ \\
&$\gamma_{d b}$  & $(0.03,0.12)$ & 0.8122 & 0.081& $2.70e^{-3}$& $6.33e^{-5}$  \\
\hline
\multirow{12}{*}{Anthrax}&$\mathrm{r}$  & $(0.002,0.004)$ & 1.016 & 0.085 & $5.18e^{-4}$ & $1.49e^{-5}$\\
&$\mu$  & $(0.001,0.002)$ & 0.6951 & 0.1524& $5.16e^{-4}$ & $1.44e^{-5}$\\
&$\kappa$  & $(0.75,1.25)$ & 2.2813 & 0.3621 & $5.21e^{-4}$ & $1.36e^{-5}$\\
&$\eta_a$  & $(0.25,0.75)$ & 0.8561 & 0.2003 & $5.19e^{-4}$ & $1.34e^{-5}$\\
&$\eta_c$  & $(0.05,0.15)$ & 2.1536 & 0.2718 & $5.16e^{-4}$ & $1.42e^{-5}$\\
&$\eta_i$  & $(0.005,0.015)$ & 3.9365 & 1.5216 & $5.28e^{-4}$ & $1.38e^{-5}$\\
&$\tau$  & $(0.05,0.15)$ &  1.1954 & 0.2008& $5.05e^{-4}$ & $1.35e^{-5}$\\
&$\gamma$  & $(0.05,0.25)$ & 1.773 & 0.616 & $5.11e^{-4}$ & $1.46e^{-5}$\\
&$\delta$  & $(0.01,0.11)$ & 2.2856 & 0.1062 & $5.31e^{-4}$ & $1.38e^{-5}$\\
&$K$  & $(75,125)$ & 0.6117 & 0.0352 & $5.26e^{-4}$ & $1.43e^{-5}$\\
&$\beta$  & $(0.001,0.003)$ & 2.9114 & 2.2214 & $5.17e^{-4}$ & $1.49e^{-5}$\\
&$\sigma$  & $(0.05,0.15)$ & 3.1815 & 1.774 & $5.22e^{-4}$ & $1.49e^{-5}$\\
\hline
\multirow{8}{*}{Polio}&$\mu$  & $(0.01,0.03)$ & 0.0051& 0.0018 & $3.76e^{-3}$ & $2.15e^{-5}$\\
&$\alpha$  & $(0.25,0.75)$ & 0.242 & 0.071& $3.55e^{-3}$ & $2.05e^{-5}$\\
&$\gamma_a$  & $(10,30)$ & 0.3915 & 0.018& $4.04e^{-3}$ & $2.25e^{-5}$ \\
&$\gamma_c$  & $(20,60)$ & 0.3714 & 0.033& $4.15e^{-3}$ & $2.13e^{-5}$ \\
&$\beta_{a a}$  & $(25,65)$ & 0.6116 & 0.0254& $3.72e^{-3}$ & $2.12e^{-5}$\\
&$\beta_{c c}$  & $(75,125)$ & 0.6115 & 0.0515& $3.85e^{-3}$ & $2.14e^{-5}$ \\
&$\beta_{a c}$  & $(-0.5,0.5)$ & 0.0003 & 0.0001 & $3.91e^{-3}$ & $2.11e^{-5}$\\
&$\beta_{c a}$  & $(-0.5,0.5)$ & 0.0003 & 0.0002 & $3.86e^{-3}$ & $2.19e^{-5}$\\
\hline
\multirow{4}{*}{Measles}&$\mu$  & $(0.01,0.03)$ & 1.514 & 0.158 & $5.51e^{-3}$&$1.14e^{-5}$  \\
&$\beta_1$  & $(0.15,0.55)$ & 2.614 & 0.723& $5.61e^{-3}$&$1.07e^{-5}$ \\
&$\gamma$ & $(75,125)$ & 2.515 & 0.365& $5.40e^{-3}$&$1.17e^{-5}$ \\
&$\sigma$ & $(20,50)$ & 3.015 & 0.176 & $5.52e^{-3}$&$1.04e^{-5}$\\
\hline
&$\mathrm{a}$  & $(0.3,0.7)$ & 0.0471 & 0.0115& $4.29e^{-3}$ & $8.02e^{-5}$ \\
&$\mathrm{~b}$  & $(0.2,0.6)$ & 0.621 & 0.153 & $4.31e^{-3}$ & $8.40e^{-5}$ \\
&$\mathrm{c}$  & $(0.3,0.7)$ & 0.1757 & 0.106 & $4.31e^{-3}$ & $8.47e^{-5}$\\
&$\eta$  & $(0.05,0.15)$ & 0.0555 & 0.012 & $4.33e^{-3}$ & $8.51e^{-5}$\\
&$\beta$  & $(0.045,0.055)$ & 0.3516 & 0.0514 & $4.21e^{-3}$ & $8.17e^{-5}$\\
&$\kappa$  & $(0.2,1.0)$ & 0.5115 & 0.0341 & $4.22e^{-3}$ & $8.30e^{-5}$\\
&$\tau$  & $(0.1,0.5)$ & 2.222 & 0.695 & $4.25e^{-3}$ & $8.35e^{-5}$\\
\multirow{5}{*}{Zika}&$\theta$  & $(10,30)$ &  0.3014 & 0.1112& $4.37e^{-3}$ & $8.37e^{-5}$\\
&$\mathrm{~m}$ & $(3,7)$ & 4.935 & 0.685& $4.24e^{-3}$ & $8.15e^{-5}$\\
&$V_b$ & $(0.175,0.225)$ & 0.1762 & 0.085 & $4.26e^{-3}$ & $8.23e^{-5}$\\
&$V_v$  & $(7.5,12.5)$ & 0.159 & 0.014 & $4.39e^{-3}$ & $8.33e^{-5}$\\
&$\gamma_{b 1}$ & $(0.1,0.3)$ & 0.4425 & 0.0319& $4.28e^{-3}$ & $8.23e^{-5}$\\
&$\gamma_{b 2}$ & $(0.03,0.07)$ & 0.358 & 0.264 & $4.26e^{-3}$ & $8.36e^{-5}$\\
&$\gamma_b$ & $(0.12,0.18)$ & 1.081 & 0.362 & $4.26e^{-3}$ & $8.41e^{-5}$\\
&$\mu_v$ & $(0.05,0.09)$ & 1.355 & 0.415 & $4.33e^{-3}$ & $8.43e^{-5}$\\
\hline
\label{tab:parameterwise_error}
\end{longtable}

\begin{figure}[h]
    \centering
    \includegraphics[width=\textwidth, height=\textheight]{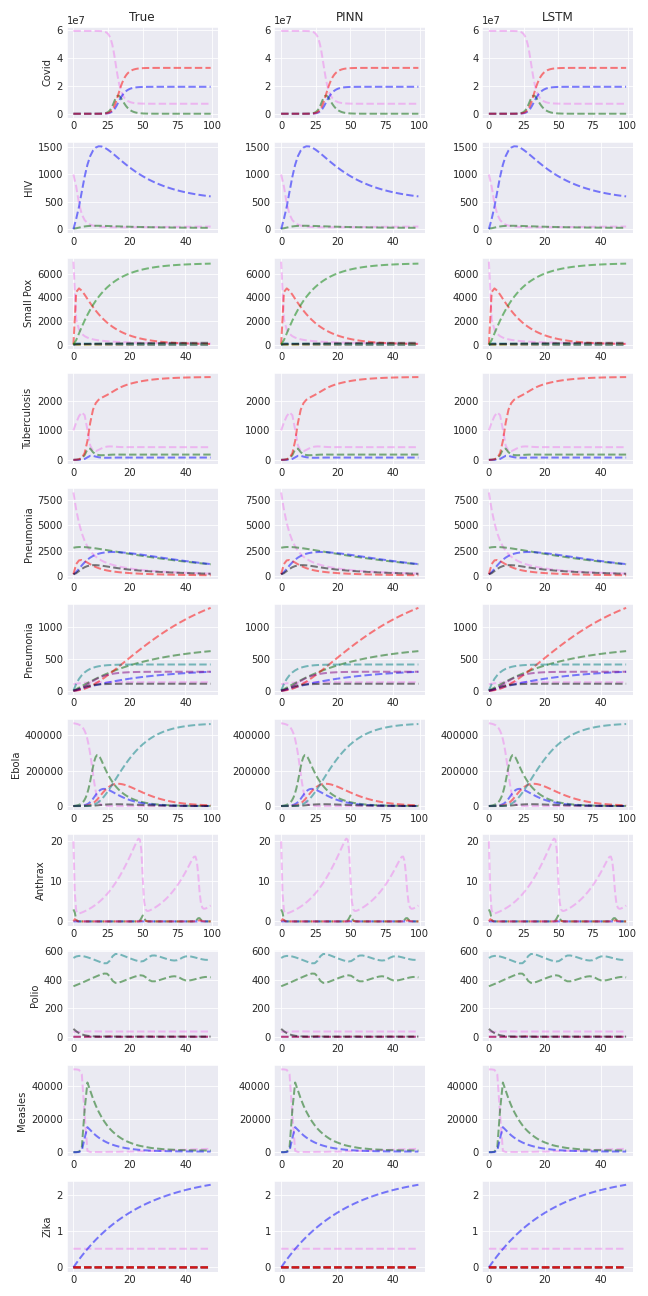}
    \caption{Comparison of True, PINN, LSTM trajectories on benchmark task}
    \label{fig:All}
\end{figure}

\end{document}

%% file: math_commands.tex
\usepackage{amsmath,amsfonts,bm}

\def\eqref#1{equation~\ref{#1}}

\def\1{\bm{1}}

\DeclareMathAlphabet{\mathsfit}{\encodingdefault}{\sfdefault}{m}{sl}
\SetMathAlphabet{\mathsfit}{bold}{\encodingdefault}{\sfdefault}{bx}{n}

